\documentclass[twocolumn]{article}

\usepackage{geometry}
\usepackage{cite}
\usepackage{caption}
\usepackage{graphicx}
\usepackage{amsmath}
\usepackage{amssymb}
\usepackage{textcomp}
\usepackage{lmodern}
\usepackage{authblk}
\usepackage{wrapfig}

\newcommand{\onlinecite}[1]{\hspace{-1 ex} \nocite{#1}\citenum{#1}} 

\begin{document}

\twocolumn[
  \begin{@twocolumnfalse}
    %\maketitle
    
\begin{center}
\LARGE Fluxonic processing of photonic synapse events\\
\vspace{0.2em}
\Large Jeffrey M. Shainline\\
\vspace{0.0em}
\textit{\small National Institute of Standards and Technology, Boulder, CO, 80305}\\
\vspace{0.3em}
\small April 1st, 2019
\end{center}

\begin{abstract}
Much of the information processing performed by a neuron occurs in the dendritic tree. For neural systems using light for communication, it is advantageous to convert signals to the electronic domain at synaptic terminals so dendritic computation can be performed with electrical circuits. Here we present circuits based on Josephson junctions and mutual inductors that act as dendrites, processing signals from synapses receiving single-photon communication events with superconducting detectors. We show simulations of circuits performing basic temporal filtering, logical operations, and nonlinear transfer functions. We further show how the synaptic signal from a single-photon can fan out locally in the electronic domain to enable the dendrites of the receiving neuron to process a photonic synapse event or pulse train in multiple different ways simultaneously. Such a technique makes efficient use of photons, energy, space, and information.
\vspace{3em}
\end{abstract}
\end{@twocolumnfalse}
]

\section{\label{sec:introduction}Introduction}
A neuron is a complex information processing device \cite{ko1997}, integrating signals from thousands of inputs and producing pulses when those signals reach threshold. These neuronal firing events consume the most energy of any operation performed by the neuron. To optimize spatial, temporal and energy efficiency, the neurons receiving the signals must extract as much information as possible from each pulse \cite{lase2003}. Neurons accomplish this through processing occurring in synapses and dendrites. Because neural information is based on temporal sequences of pulses, the relevant processing involves applying temporal and logical filters to extract relevant data. For example, synapses perform temporal filtering of pulse trains to identify rising edges and to identify pulse trains exceeding some duration or number of pulses \cite{abre2004}. Dendrites receive and further process synaptic signals. The operations performed by dendrites include leaky integration \cite{geki2002}; logical operations \cite{stsp2015}; identification of coincidences \cite{stsp2015} and sequences \cite{stse2007,haah2015} between synapses from different neurons; and nonlinear thresholding transfer functions on signals from groups of synapses \cite{sava2017}. Inhibitory neurons in the network can temporarily suppress the activity of a dendrite to dynamically direct attention to information of interest \cite{enfr2001}, thereby adapting the structural network into myriad functional networks \cite{brme2010}.

Within a point-neuron model \cite{geki2002}, each neuron performs leaky integration of the synaptic activities with a single decay time constant, $\tau$. Thus, a neuron is capable of answering the question, ``Is the sum of activity across all synapses in the last $\tau$ seconds greater than threshold?'' If the answer is yes, the neuron produces a pulse. While such a model may be useful for certain neuromorphic computations, it assumes that each neuron ignores or is incapable of utilizing nearly all the information to which it has access. In this work we develop circuits that will enable a neuron to answer subtle and varied questions such as, ``How long has it been since neuron $i$ last produced a pulse?'' ``How many pulse trains have begun and then ceased on neuron $i$ in the last $\tau_i$ seconds?'' ``How many times have neurons $i$ and $j$ fired within $\tau_{ij}$ seconds of each other in the last $\tau_q$ seconds?'' ``Have five or more of the neurons in cluster $x$ fired in the last $\tau_x$ seconds?''

For hardware to be efficient for neural information processing, synaptic and dendritic operations must be efficiently manifest in constituent devices. We have argued elsewhere that light is promising for communication in neural systems because it enables the fan-out and energy efficiency necessary for large neural systems, and that utilization of superconducting single-photon detectors enables communication at the lowest possible light levels \cite{shbu2017}. Subsequent work considered specific synaptic and neuronal circuits suitable for point-neuron behavior, introducing circuits capable of transducing single-photon communication events to the electronic domain for subsequent information processing \cite{sh2018,sh2018_full}. References \onlinecite{sh2018} and \onlinecite{sh2018_full} discussed basic synaptic functionality, plasticity, neuronal integration, thresholding, and the production of light during a neuronal firing event\textemdash all functions necessary for point neurons implemented with superconducting optoelectronic hardware. Due to the prominent role of flux storage loops, these circuits are referred to as loop neurons. The significance of light and superconductors for scaling was analyzed in Ref.\,\onlinecite{sh2018_ICRC}. Other work has investigated photonics \cite{nash2013,tafe2017,prsh2017,shha2016,chsa2018} and superconducting electronics \cite{hias2007,crsc2010,ru2016,sele2017,scdo2018,kafu2018} for neuromorphic computation, but to our knowledge none of this work has pursued dendritic processing beyond point neurons or the integration of photonics with superconducting electronics to leverage their complementary strengths for communication and computation. 

\begin{figure}[tb]
    \centering{\includegraphics[width=8.6cm]{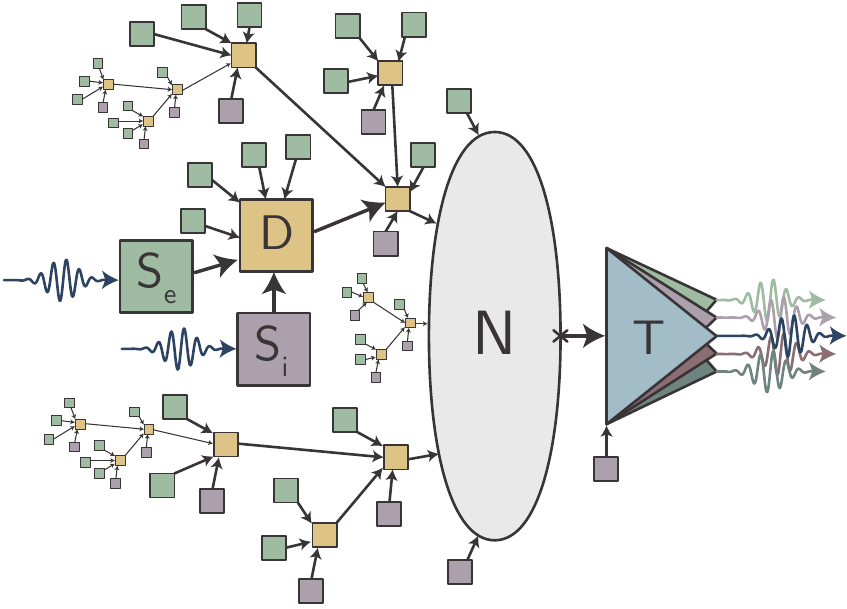}}
	\captionof{figure}{\label{fig:schematic}Schematic of the neuron under consideration. The complex structure consists of excitatory and inhibitory synapses ($\mathsf{S_e}$ and $\mathsf{S_i}$) that feed into dendrites ($\mathsf{D}$). Each dendrite performs computations on the inputs and communicates the result to other dendrites for further processing or on to the cell body of the neuron ($\mathsf{N}$). The neuron itself acts as the final thresholding stage, and when its threshold is reached, light is produced by the transmitter ($\mathsf{T}$), which is routed to downstream synaptic connections.}
\end{figure}
\begin{figure}%[tb]
    \centering{\includegraphics[width=8.6cm]{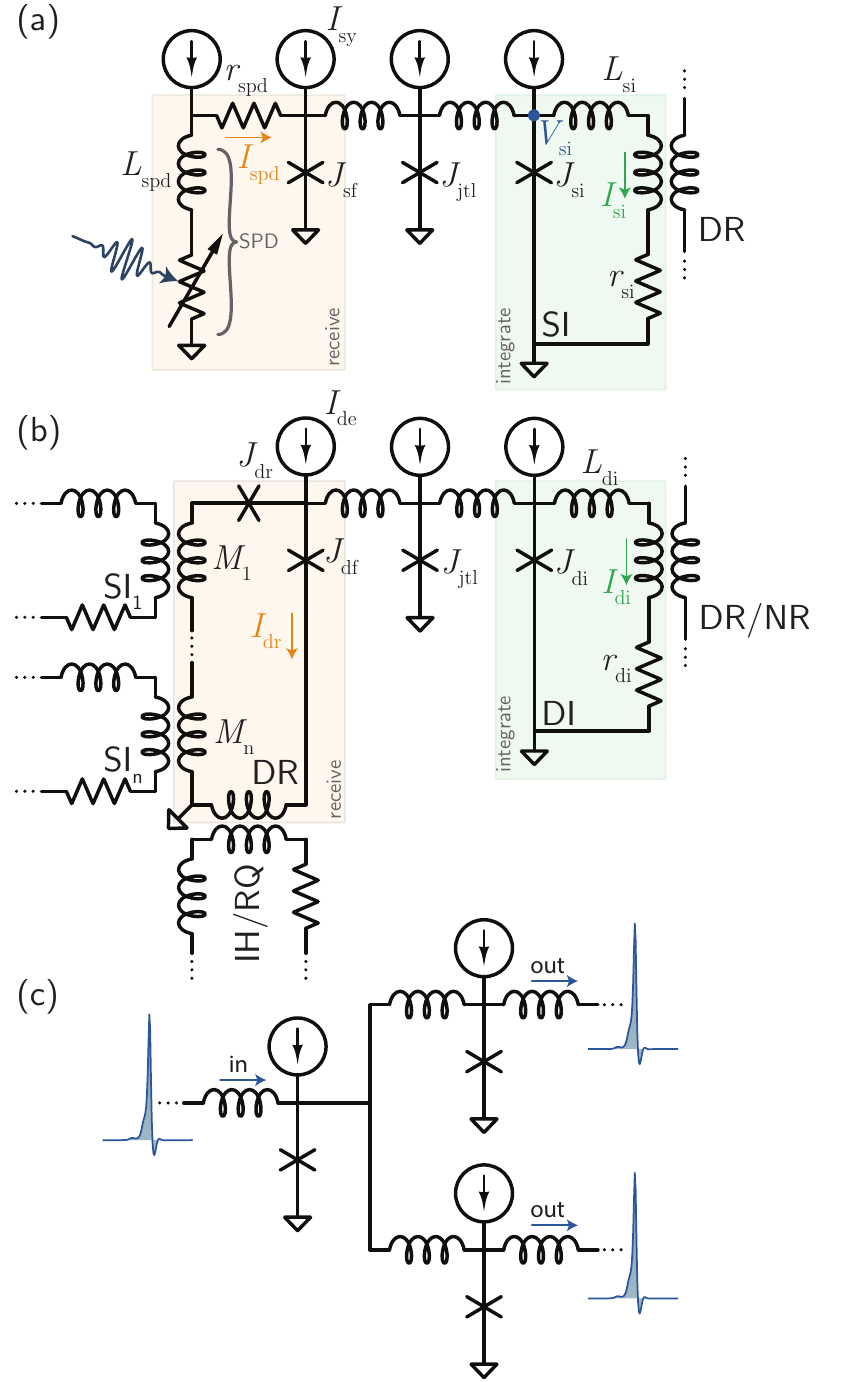}}
	\captionof{figure}{\label{fig:circuits}Diagrams of the circuits under consideration. (a) Synaptic transducer. A photonic communication event with one or more photons diverts current from the single-photon detector (SPD) to the synaptic firing junction ($I_{\mathrm{spd}}$ to $J_{\mathrm{sf}}$). A series of fluxons is produced, and these fluxons traverse the Josephson transmission line ($J_{\mathrm{jtl}}$) and result in an integrated current in the synaptic integration loop ($I_{\mathrm{si}}$). This synaptic signal is communicated to a dendritic receiving (DR) loop through a mutual inductor. (b) Dendritic circuit. The dendritic receiver loop sums the signals from afferent synapses, and upon reaching the threshold established by the dendritic firing junction ($J_{\mathrm{df}}$), one or a series of fluxons is generated. Inhibitory (IH) or rapid query (RQ) synapses can also be established on the DR loop. The generated signal is communicated to other dendritic receiving loops or to the receiving loop of the neuron cell body. While drawn in the same place, either IH or RQ will be present on a given dendrite, and these loops require opposite signs of mutual inductance. (c) Fluxon pulse splitter. This circuit is used to make electronic copies of the information generated by a synapse or a dendrite, and the amplitudes of the current pulses at the outputs are restored to the input level.}
\end{figure}

The purpose of this paper is to consider specific circuits implementing a more elaborate model for superconducting optoelectronic neural information processing in which the dendritic tree extracts significantly more information about synaptic activities than a simple sliding average. The model is illustrated schematically in Fig.\,\ref{fig:schematic}. We use the term ``dendritic tree'' to refer to a neuron's input synapses and dendrites collectively, and Fig.\,\ref{fig:schematic} is intended to illustrate the potential complexity of the dendritic tree. We discuss three elemental circuits (Fig.\,\ref{fig:circuits}) that can be used as building blocks to perform many synaptic and dendritic functions. These functions include leaky integration, temporal filtering of afferent pulse trains, logical operations, detection of coincidences between activities of input neurons, inhibition, and power-law memory retention of synaptic activity. In biological systems, these functions occur through nonlinearities resulting from dendritic conductances and arbor morphology \cite{stse2007,stsp2015}. The Josephson circuits presented here are not intended to quantitatively reproduce biological behaviors, but rather to perform logical, temporal, and nonlinear functions in the spirit of synaptic and dendritic processing. Josephson circuits are remarkably capable of these operations due to the nonlinearity established by the existence of a critical current; the avoidance of cross talk and current leakage pathways enabled by coupling through mutual inductors; and the ability to establish essentially arbitrary time constants across many orders of magnitude by choosing the inductance and resistance of current storage loops. 

This work is based on time-domain circuit simulations of the three elemental circuits shown in Fig.\,\ref{fig:circuits} when arranged in various configurations. In Sec.\,\ref{sec:synapse} we review the basic operations of a synapse that transduces a single-photon communication event to the superconducting electronic domain for information processing, and in Sec.\,\ref{sec:short_term} we consider operations performed on pulse trains at a single synapse, usually associated with short-term plasticity and synaptic computation. In Sec.\,\ref{sec:correlations} we consider the detection of coincidences between two or more synapses, and we show how the same circuits can be used with broken temporal symmetry to identify sequences of activity. For these various fragments of information to be utilized only when relevant, inhibition can be used to silence specific dendrites at appropriate times, as discussed in Sec.\,\ref{sec:inhibition_and_rapid_query}. A central premise of the work in Refs.\,\onlinecite{shbu2017,sh2018,sh2018_full,sh2018_ICRC} is that scalable neural systems will benefit from the fan-out and efficiency of few-photon communication. Yet when superconducting electronic circuits are employed for computation, even few-photon communication events represent a significant energy expense. In Sec.\,\ref{sec:fluxonic_fanout} we discuss the use of superconducting splitters to make copies of photonic synapse events so that answers to all of the questions listed above can be simultaneously present in the dendritic tree through processing of the signal from a single photon. Section \ref{sec:discussion} contains a discussion of the results.

\section{\label{sec:synapse}Photon-to-fluxon transduction at a synapse}
Analysis of fluxonic processing of photonic synapse events begins with consideration of the circuit that transduces a single-photon detection event to the superconducting electronic domain in the form of a series of fluxons. The circuit that accomplishes this is shown in Fig.\,\ref{fig:circuits}(a). This circuit was first introduced in Ref.\,\onlinecite{sh2018} and described in more detail in Ref.\,\onlinecite{sh2018_full}. The circuit comprises an initial receiver/transducer section, consisting of a superconducting-nanowire single-photon detector (SPD) \cite{gook2001,nata2012,liyo2013,mave2013} in parallel with a Josephson junction (JJ) \cite{ti1996,vatu1998,ka1999}. In the steady state, the SPD (drawn as a variable resistor in series with an inductor) has zero resistance, and thus its entire bias current flows directly through it to ground. The synaptic firing junction, $J_{\mathrm{sf}}$, is biased below its critical current ($I_{\mathrm{c}}$) by the synaptic bias current, $I_{\mathrm{sy}}$. Upon absorption of a photon, the variable resistor of the SPD switches temporarily to a high-resistance state ($5$\,k$\Omega$) for a short duration ($200$\,ps) \cite{yake2007}. The current through the SPD is diverted across a resistor ($I_{\mathrm{spd}}$ across $r_{\mathrm{spd}}$ in Fig.\,\ref{fig:circuits}(a)) and to $J_{\mathrm{sf}}$. At this point, the sum of the currents across $J_{\mathrm{sf}}$ exceeds $I_{\mathrm{c}}$, and the junction produces a series of fluxons \cite{ti1996,vatu1998,ka1999}. These fluxons propagate along the Josephson transmission line \cite{vatu1998,ka1999}, and are stored in the synaptic integration (SI) loop. The Josephson transmission line simply serves to isolate the activity of the receiver portion of the circuit from the integration loop, allowing their circuit parameters to be optimized independently. After the 200\,ps photon detection event, the bias current returns to the SPD with the time constant of $\tau_{\mathrm{spd}} = L_{\mathrm{spd}}/r_{\mathrm{spd}}$. This time constant has a minimum functional value determined by the electro-thermal properties of the nanowire \cite{yake2007}, and throughout this work this time constant is fixed at $\tau_{\mathrm{si}} = 10$\,ns, and the bias to the SPD is fixed at 10\,\textmu A. The number of fluxons created during a synaptic firing event depends on the net current across $J_{\mathrm{sf}}$ as well as the duration during which $J_{\mathrm{sf}}$ is biased above $I_{\mathrm{c}}$. With $\tau_{\mathrm{si}}$ and the bias to the SPD fixed, the number of fluxons, and thus the synaptic weight, are dynamically adaptable by changing the synaptic bias current, $I_{\mathrm{sy}}$. More details regarding $I_{\mathrm{sy}}$ and the associated plasticity mechanisms are given in Ref.\,\onlinecite{sh2018_full}.
 
\begin{figure}[tb]
    \centering{\includegraphics[width=8.6cm]{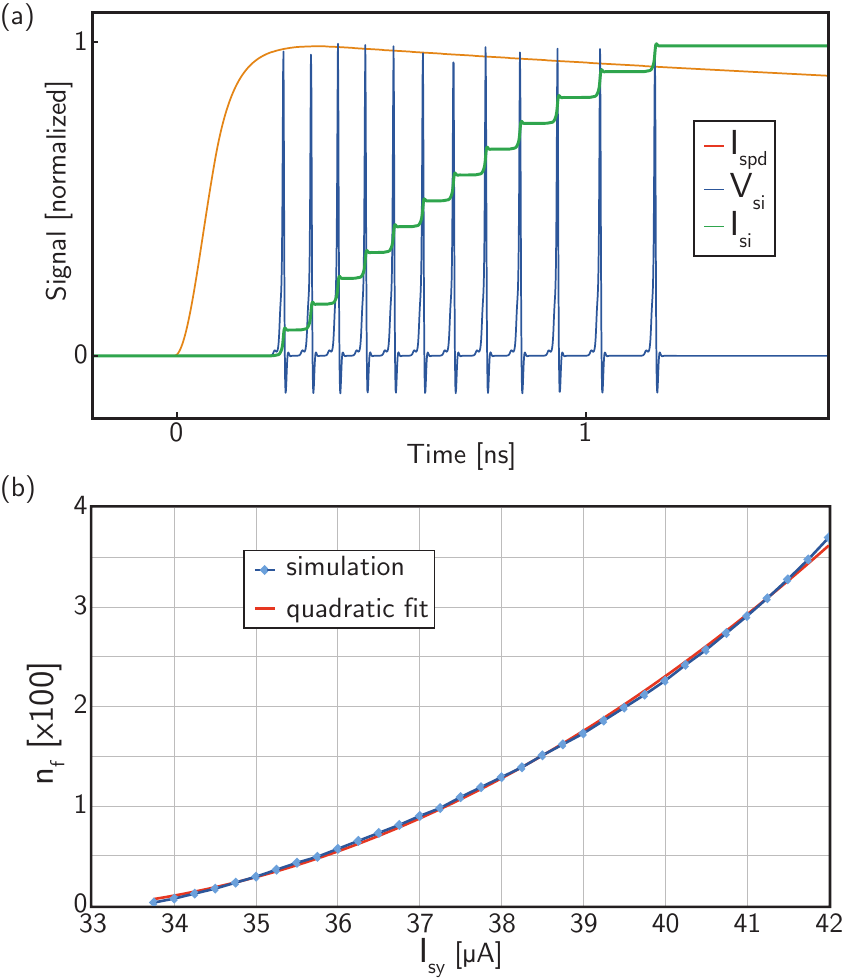}}
	\captionof{figure}{\label{fig:sffg_basic}Basic operation of the photon-to-fluxon synaptic transducer. (a) Temporal activity of the circuit in Fig.\,\ref{fig:circuits}(a) during a synaptic firing event. The traces are color-coded with the currents and voltages labeled in Fig.\,\ref{fig:circuits}(a), and all traces have been independently normalized for display on the same plot. (b) Demonstration of variable synaptic weight. The number of fluxons generated during a synaptic firing event ($n_{\mathrm{f}}$) is plotted as a function of the synaptic bias current ($I_{\mathrm{sy}}$). A fit to a second-order polynomial is also shown.}
\end{figure}
The temporal activity of the circuit in Fig.\,\ref{fig:circuits}(a) during a synaptic firing event is shown in Fig.\,\ref{fig:sffg_basic}(a). Throughout this work, WRSpice \cite{wh1991} has been used to simulate all circuits. All JJs have $I_{\mathrm{c}} = 40$\,\textmu A and $\beta_{\mathrm{c}} = 0.95$. The yellow trace in Fig.\,\ref{fig:sffg_basic}(a) shows the current diverted from the SPD after a photon has been received. The blue trace shows the voltage pulses as the fluxons enter the SI loop. As each fluxon enters the loop, it introduces a discrete, fixed value of current given by $I_{\phi} = \Phi_0/L_{\mathrm{si}}$, where $\Phi_0 \approx 2\times 10^{-15}$\,Wb is the magnetic flux quantum, and $L_{\mathrm{si}}$ is the inductance of the synaptic integration loop. We assume the value of $L_{\mathrm{si}}$ is chosen by design independently for each synapse and set in hardware at the time of fabrication. The green trace in Fig.\,\ref{fig:sffg_basic}(a) shows the increase in current as the fluxons enter the SI loop during a synaptic firing event. The discrete steps with each fluxon are evident, and the total amount of current added to the SI loop during a synaptic firing event depends on both the number of fluxons generated during the firing event (controlled dynamically by $I_{\mathrm{sy}}$) and the inductance of the SI loop (set in hardware as $L_{\mathrm{si}}$). 

The role of $I_{\mathrm{sy}}$ is to adapt the synaptic weight by changing the number of fluxons generated during a synaptic firing event. In Fig.\,\ref{fig:sffg_basic}(b) we show the number of fluxons generated during a synaptic firing event as a function of $I_{\mathrm{sy}}$. The fit shows close agreement with a quadratic function. This method of establishing and adapting the synaptic weight has several important properties. First, it is slowly varying, so small changes in $I_{\mathrm{sy}}$ result in small changes in the synaptic efficacy. Second, the function is monotonic, so increases in $I_{\mathrm{sy}}$ always result in increased synaptic efficacy, while decreases in $I_{\mathrm{sy}}$ always result in decreases in synaptic efficacy. This is necessary to enable activity-based plasticity mechanisms \cite{somi2000,mage2012}, which have been explored in the context of these circuits in Ref.\,\onlinecite{sh2018_full}. Third, the bias $I_{\mathrm{sy}}$ can be bounded so synaptic strength never exceeds a certain limit, and runaway activity is not possible. Finally, the integer number of fluxons generated can be made to cover a broad range so that analog synapses of relatively high bit depth can be achieved. Figure \ref{fig:sffg_basic}(b) shows that over eight bits (256 levels) can be utilized, and throughout this work we find the range of eight to 10 bits to be a comfortable working range for the circuits under consideration. This is much lower than the 64-bit processors used for high-arithmetic-depth numerical calculations. Yet neural computation benefits from performing lower-resolution operations with high efficiency and accuracy gained through redundancy and parallelism. 

After a photonic communication event has been detected, the synaptic weight has been set as the number of fluxons created, and current has been added to the SI loop, further processing ensues. The electrical current generated by the synapse event can be stored for a chosen amount of time. This is determined by the leak rate of the SI loop, selected by design and set in hardware with the time constant $\tau_{\mathrm{si}} = L_{\mathrm{si}}/r_{\mathrm{si}}$. Note that $\tau_{\mathrm{si}}$ is entirely independent of $\tau_{\mathrm{spd}}$, and because we consider superconducting circuits, memory of a synaptic event can persist indefinitely. Also note that while the amount of current added to the SI loop during a synaptic firing event depends on $L_{\mathrm{si}}$, $r_{\mathrm{si}}$ can be chosen independently from $L_{\mathrm{si}}$, thereby enabling the amount of current and its storage time to be separately selected. The current can be released quickly, on the order of the SPD reset time of 10\,ns, or it can be stored 10 or 100 times longer to retain a memory of the event for as long as required. In this work we mainly consider decay times spanning two orders of magnitude, from 10\,ns to 1\,\textmu s. 

In biological neural systems, processing among local clusters of neurons occurs primarily through fast activity in the range of gamma frequencies (30\,Hz - 80\,Hz) \cite{budr2004,bu2006}. This frequency range emerges because it reaches the upper limit of speed for the excitatory pyramidal neurons participating in the activity. In the superconducting optoelectronic hardware under consideration, this upper speed limit is in the tens of megahertz, limited by the reset time of the SPDs in the synapses and of the transmitter circuits that generate neuronal firing events \cite{sh2018_full}. Here we take the upper firing rate to be 100\,MHz for numerical simplicity. Therefore, we expect the neurons under consideration to demonstrate behavior similar to gamma oscillations, bursting with inter-spike intervals on the order of 10\,ns. Similarly, biological neural systems process information across the network as a whole through slower activity at theta frequencies (4\,Hz - 8\,Hz) \cite{budr2004,bu2006}. Mapping this scaling onto the system under consideration, we pay particular attention to gamma oscillations occurring at 100\,MHz as well as theta oscillations occurring at 10\,MHz. It is for this reason that we consider $\tau_{\mathrm{si}}$ ranging from 10\,ns to 1\,\textmu s and spike trains in the 50\,MHz to 100\,MHz range. 

\begin{figure}[tb] 
    \centering{\includegraphics[width=8.6cm]{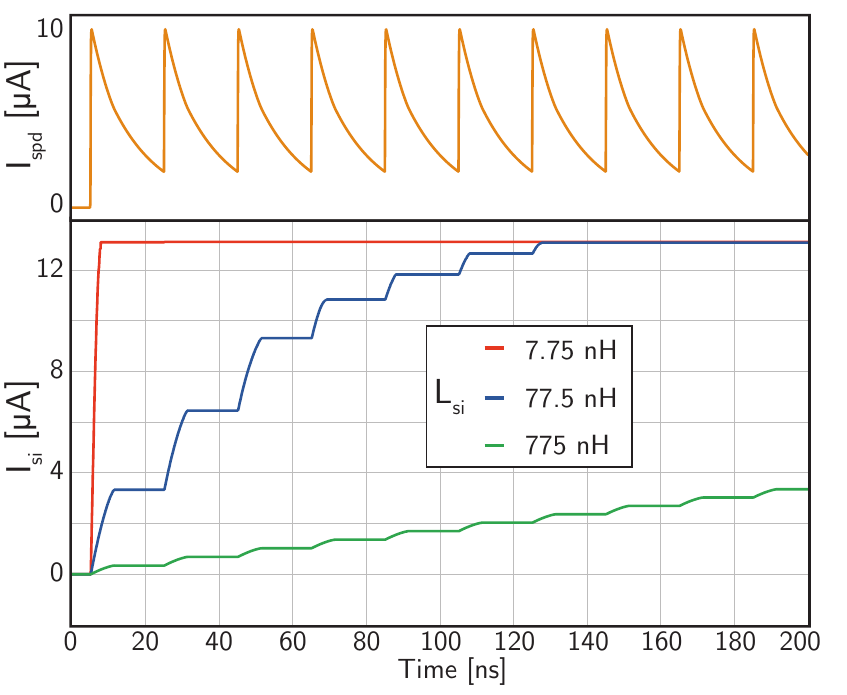}}
	\captionof{figure}{\label{fig:sffg_betaL_tauSi}Filling of the SI loop in response to a pulse train. The upper panel shows the current pulses generated by detection events at the SPD. The lower panel shows time traces of $I_{\mathrm{si}}$. Here $\tau_{\mathrm{si}} = \infty$ to focus attention on the manner in which the SI loop fills with current, rather than how it decays. A loop with small inductance ($L_{\mathrm{si}} = 7.75$\,nH) will saturate after a single photon detection event, while a loop with large inductance ($L_{\mathrm{si}} = 775$\,nH) can store the signals from many synaptic firing events.}
\end{figure}
In addition to signal decay from a synaptic integration loop, we must also consider saturation, as shown in Fig.\,\ref{fig:sffg_betaL_tauSi}. As stated above, the current associated with a fluxon being generated in a loop of inductance $L$ is $I_{\phi} = \Phi_0/L$. This current circulates in the direction opposing the applied bias to the JJ. The number of fluxons that can enter the loop before the cumulative opposing bias equals $I_{\mathrm{c}}$ is given by $I_{\mathrm{c}}/I_{\phi} = L I_{\mathrm{c}}/\Phi_0 = \beta_{\mathrm{L}}/2 \pi$, where $\beta_{\mathrm{L}}$ is a common parameter quantifying the flux storage capacity of a superconducting loop. $\beta_{\mathrm{L}}/2\pi$ gives an estimate for how many fluxons a given SI loop will be able to store before saturation, and the exact number also depends on the applied bias. In Fig.\,\ref{fig:sffg_betaL_tauSi} we show the integrated current in an SI loop as a function of time in response to a periodic train of pulses with 20\,ns inter-spike interval. Here we fix $\tau_{\mathrm{si}} = \infty$ and vary the inductance of the loop. In these simulations, the value of $I_{\mathrm{sy}}$ was fixed at 38\,\textmu A, so 129 flux quanta ($> 2^7$) are generated during each synaptic firing event until the loop nears saturation, at which point the effective synaptic weight is suppressed, demonstrating a simple form of short-term plasticity. With a small value of $L_{\mathrm{si}}$, the quantity $\beta_{\mathrm{L}}/2\pi = L_{\mathrm{si}} I_{\mathrm{c}}/\Phi_0 =  150$, and the loop saturates after a single synaptic firing event. With an intermediate value of $L_{\mathrm{si}} = 77.5$\,nH, $\beta_{\mathrm{L}}/2\pi = 1.5\times 10^3$, and seven synaptic firing events fill the loop. With a large value of $L_{\mathrm{si}} = 775$\,nH, $\beta_{\mathrm{L}}/2\pi = 1.5\times 10^{4}$, and the loop can hold the activity from nearly 100 synaptic firing events with this value of $I_{\mathrm{sy}}$. All these values of inductance are straightforward to achieve with high-kinetic-inductance materials. Note that in digital superconducting electronics $\beta_{\mathrm{L}}/2\pi = 1.5$, so a loop can hold a single fluxon to represent a bit. Figure \ref{fig:sffg_betaL_tauSi} shows the control one has in design over the capacity of the SI loop. The loop can operate as a binary device switching from a low to high state with each synapse event, or it can act as an analog device capable of representing many synapse events with distinct values of current. This saturation is a simple form of nonlinearity present in the synapse. 

As we have described, the two basic degrees of freedom of the SI loop are the signal storage time and storage capacity. We now proceed to explore the use of such synapses to extract information from pulse trains.

\section{\label{sec:short_term}Operations on pulse trains at a single synapse}
\begin{figure}[tb] 
    \centering{\includegraphics[width=8.6cm]{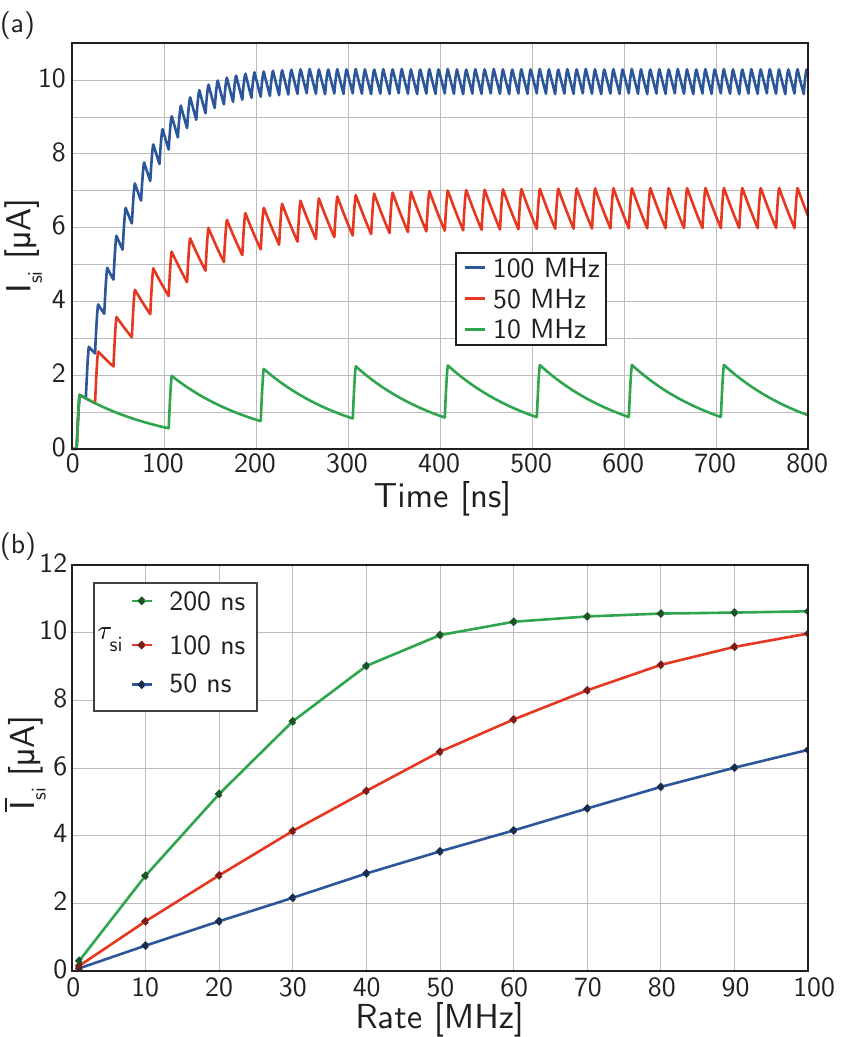}}
	\captionof{figure}{\label{fig:sffg_rate_conversion}Rate-to-current conversion at a synaptic transducer. (a) $I_{\mathrm{si}}$ as a function of time as pulse trains of various frequencies are incident upon the synapse. (b) Systematic analysis of rate-to-current mapping for SI loops of three decay time constants. To obtain these curves, temporal traces like those of (a) have been analyzed once the steady state has been reached. Each data point in (b) results from time averaging a trace such as those in (a) over a single interspike interval: $\bar{I}_{\mathrm{si}} = (t_2-t_1)^{-1}\int_{t_1}^{t_2} I_{\mathrm{si}}(t) dt$, where $t_1$ and $t_2$ are arrival times of consecutive photons at the synapse after the steady state has been reached.}
\end{figure}
As an example of one form of processing that can be performed using the synaptic circuit of Fig.\,\ref{fig:circuits}(a), Fig.\,\ref{fig:sffg_rate_conversion} considers the operation of rate-to-current conversion. The first term of the Volterra expansion of a spike train corresponds to the time-averaged spike rate \cite{geki2002}, so a neuron must be able to decode this information. This can be accomplished with the synaptic transducer of Fig.\,\ref{fig:circuits}(a) when the SI loop is given a leak rate, as discussed above. The circuit behaves as a standard leaky integrator modeled as $\dot{I}_{\mathrm{si}} = \alpha-I_{\mathrm{si}}/\tau_{\mathrm{si}}$, where $\alpha$ is the rate of current added to the SI loop by synaptic firing events. The leaky integrator model has the steady-state solution $I_{\mathrm{si}} = \alpha \tau_{\mathrm{si}}$, indicating that the current in the loop is proportional to the rate of input spikes. In Fig.\,\ref{fig:sffg_rate_conversion}(a) we show temporal traces of the current $I_{\mathrm{si}}$ in the presence of afferent activity at various rates for a loop with $\tau_{\mathrm{si}} = $100\,ns and $L_{\mathrm{si}} = $77.5\,nH, and it can be seen that the time-averaged value of $I_{\mathrm{si}}$ reaches steady-state. In Fig.\,\ref{fig:sffg_rate_conversion}(b) we show the time-averaged current, $\bar{I}_{\mathrm{si}}$, as a function of the synaptic firing rate for three values of $\tau_{\mathrm{si}}$. With the value $\tau_{\mathrm{si}} =$\,50\,ns, the response is linear across the entire range of gamma and theta frequencies. Linear rate-to-current conversion holds as long as the integration time of the loop is short enough to avoid saturation, that is, $\alpha \tau_{\mathrm{si}} < I_{\mathrm{si}}^{\mathrm{sat}}$. With $\tau_{\mathrm{si}} = 200$\,ns, the loop reaches saturation, and higher input frequencies do not code unique information. If linear operation is desired, one must choose the time constant of the loop to be commensurate with the frequencies to be detected, or if nonlinear saturation is desired, longer integration times can be utilized. If increased dynamic range is advantageous, one can utilize the splitter of Fig.\,\ref{fig:circuits}(c) to activate multiple SI loops with different time constants from the same photonic synapse, as described in Sec.\,\ref{sec:fluxonic_fanout}.

The synaptic transducer and SI loop of Fig.\,\ref{fig:circuits}(a) on its own can achieve straightforward rate-to-current conversion to make use of rate-coded neuronal information. Yet when $I_{\mathrm{si}}$ is coupled to the circuit of Fig.\,\ref{fig:circuits}(b) through a mutual inductor, significantly more functionality can be achieved, as we will discuss shortly. Let us first describe the basic operation of the circuit in Fig.\,\ref{fig:circuits}(b), which we refer to as a dendritic processing circuit or dendrite. The dendritic processing circuit of Fig.\,\ref{fig:circuits}(b) is similar to the synaptic transducer circuit of Fig.\,\ref{fig:circuits}(a). Unlike the synapse, which receives photonic input, the dendrite receives input as flux coupled through mutual inductors. In the steady state, all junctions are biased below $I_{\mathrm{c}}$. Afferent input to the dendritic receiving (DR) loop from one or more SI loops increases the bias to the dendritic firing junction ($J_{\mathrm{df}}$). When the net bias to $J_{\mathrm{df}}$ exceeds $I_{\mathrm{c}}$, one or more fluxons will be produced, they will traverse the JTL, and they will add flux to the dendritic integration (DI) loop, just as in the case of the synapse. The role of the dendritic reset junction ($J_{\mathrm{dr}}$) is to release the flux generated by $J_{\mathrm{df}}$ from the DR loop, thereby resetting the loop to the state prior to firing.

The use of mutual inductors is advantageous for coupling multiple synapses to a single dendrite because mutual inductors reduce cross talk between synapses to a very low level. In general, SI loops have a self-inductance of at least 1\,nH, and possibly up to 10\,\textmu H. The mutual inductors considered here are asymmetric with the inductor in the SI loop being on the order of 100\,pH and the coupled inductor in the DR loop being on the order of 10\,pH. The total inductance of the DR loop is on the order of 100\,pH. Thus, when current is circulating in one SI loop, appreciable current is coupled to the DR loop, while the parasitic current coupled into other SI loops is significantly smaller. Using typical numbers from the circuits studied in this work, the parasitic current coupled to an adjacent SI loop is roughly one thousandth the current induced in the DR loop, with $I_{\mathrm{dr}}$ being on the order of microamps.

The dendritic circuit under consideration is reminiscent of a DC SQUID \cite{vatu1998,ka1999}, and is also similar to the neuron circuit presented in Ref.\,\onlinecite{crsc2010}. Both the synaptic and dendritic circuits explored here are similar in principle to a wide class of particle and field detectors leveraging superconducting circuits \cite{vatu1998}. The main computational attributes of the dendrite come from the biasing conditions and interplay between $J_{\mathrm{df}}$ and $J_{\mathrm{dr}}$. If the biases are established such that when $J_{\mathrm{df}}$ produces a fluxon, the current added to $J_{\mathrm{dr}}$ is insufficient to switch $J_{\mathrm{dr}}$ until the added biases from the SI loop(s) decay, the device acts like a DC-to-SFQ converter \cite{vatu1998,ka1999}. $J_{\mathrm{df}}$ will produce exactly one fluxon, and the DR loop will then be inactivated until the counter bias across $J_{\mathrm{dr}}$ due to the SI loop(s) decays, at which point $J_{\mathrm{dr}}$ will produce a fluxon countering the one produced by $J_{\mathrm{df}}$, and the loop will be reset. In this configuration, the dendritic receiver has a binary character.

The circuit can also operate in an analog mode, wherein the dendrite can produce a continuous stream of fluxons, much like the synaptic transducer. To achieve this operation, $J_{\mathrm{dr}}$ is biased closer to $I_{\mathrm{c}}$ so that a fluxon generated by $J_{\mathrm{df}}$ is sufficient to switch $J_{\mathrm{dr}}$. Thus, each time $J_{\mathrm{df}}$ produces a fluxon, it is rapidly canceled by $J_{\mathrm{dr}}$, and the DR loop is reset with no net flux. $J_{\mathrm{df}}$ will continue to produce fluxons as long as it is held above $I_{\mathrm{c}}$, and in the presence of synaptic activation (current in one or more SI loops), a stream of fluxons will be generated by $J_{\mathrm{df}}$ and stored in the DI loop. This stream may contain a large number of fluxons until the DI loop saturates, so we consider this an analog mode of operation.

Whether operating in binary or analog, the effect of the dendrite is to perform a nonlinear transfer function on its inputs and provide the output signal to the DI loop in the form of supercurrent. Just as in the SI loop, the DI loop can be configured to saturate rapidly (small $\beta_{\mathrm{L}}$) or store the signal from many threshold events (large $\beta_{\mathrm{L}}$), and the loop can be configured with a decay time constant ($\tau_{\mathrm{di}} = L_{\mathrm{di}} / r_{\mathrm{di}}$) spanning a broad range, from time scales shorter than a gamma interspike interval to as long as superconductivity can be maintained. With these basic operating principles in mind, we proceed to consider examples of dendritic processing with this circuit.

We first consider operations usually associated with synaptic computation \cite{abre2004}, namely short-term-facilitating and short-term-depressing plasticity. Some synapses are observed to provide no response or very weak response to the first pulse of a train, with the efficacy of the synapse increasing as the pulse train proceeds. This behavior is referred to as short-term-facilitating plasticity, and it can be due to dynamics within the synapse itself or to the conductance properties of a dendrite or series of dendritic compartments. Here we simulate analogous behavior with a single synaptic transducer (Fig.\,\ref{fig:circuits}(a)) coupled to a single dendritic processing circuit (Fig.\,\ref{fig:circuits}(b)). 

\begin{figure}[tb] 
    \centering{\includegraphics[width=8.6cm]{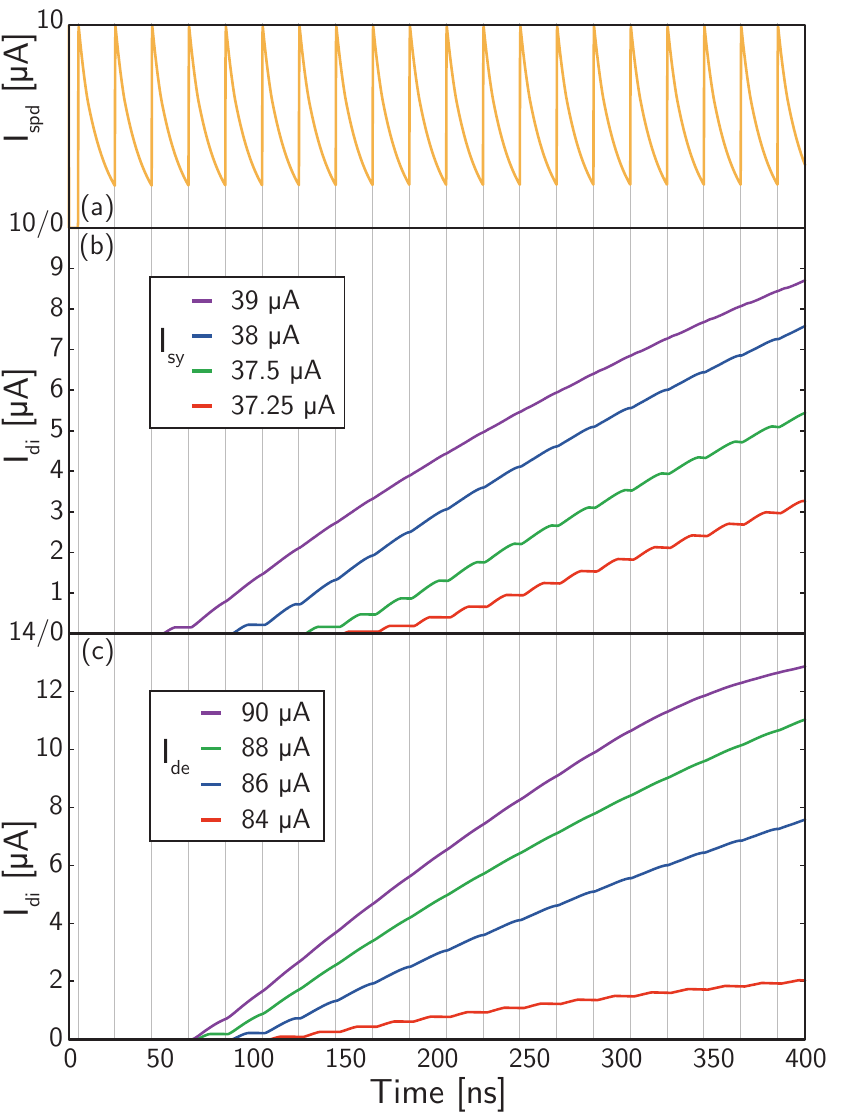}}
	\captionof{figure}{\label{fig:si_di_facilitating}Short-term-facilitating plasticity. With a single synapse coupled to a dendrite, a nonlinearity can be induced wherein multiple synaptic firing events are required to generate a signal. (a) The afferent pulse train. (b) The current in the DI loop, $I_{\mathrm{di}}$, as a function of time for several values of the synaptic bias current, $I_{\mathrm{sy}}$. (c) $I_{\mathrm{di}}$ for several values of the dendritic bias current, $I_{\mathrm{de}}$. The blue curve is the same in (b) and (c). The vertical lines spanning (a)-(c) represent the times of the synaptic firing events.}
\end{figure}
To achieve short-term-facilitating plasticity, we design an SI loop that can store the signals from multiple synaptic firing events before saturation, and we bias $J_{\mathrm{df}}$ so that the additional current induced by the first few synaptic firing events does not push the junction over $I_{\mathrm{c}}$, but after multiple synaptic firing events, $I_{\mathrm{c}}$ is exceeded and flux is added to the DI loop. We design the dendrite in analog mode for this behavior. Circuit simulations of short-term-facilitating plasticity are shown in Fig.\,\ref{fig:si_di_facilitating}. Figure \ref{fig:si_di_facilitating}(a) shows the afferent pulse train. The first pulse occurs at 5\,ns, and the interspike interval is 20\,ns. Figures \ref{fig:si_di_facilitating}(b) and (c) show the accumulated current in the DI loop as a function of time. In Fig.\,\ref{fig:si_di_facilitating}(b) the effect of the synaptic bias current, $I_{\mathrm{sy}}$ is shown. The primary effect of the dynamically reconfigurable bias current is to shift the curve left or right. With a stronger synaptic weight, more current will be added to the SI loop with each synaptic firing event, and therefore more current will be induced by the mutual inductor into the DR loop. Thus, fewer synaptic firing events are required to reach threshold in the dendritic compartment. In this example, $I_{\mathrm{sy}}$ can shift the threshold from three to eight synaptic firing events. In Fig.\,\ref{fig:si_di_facilitating}(c), the synaptic bias current is fixed at 38\,\textmu A, while the dendritic bias current, $I_{\mathrm{de}}$, is varied. Change in $I_{\mathrm{de}}$ has less of an effect on the number of pulses required to reach threshold, but it significantly affects the number of fluxons generated by $J_{\mathrm{df}}$ each time a synaptic firing event occurs, which is related to the slope of the traces in Fig.\,\ref{fig:si_di_facilitating}(c). The effect of the dendritic bias current, $I_{\mathrm{de}}$, is therefore analogous to the effect of the synaptic bias current, $I_{\mathrm{sy}}$. We therefore anticipate that $I_{\mathrm{de}}$ will provide a dynamically reconfigurable circuit parameter that can be used to establish a ``dendritic weight'' and can be used for long-term plasticity and learning. 

\begin{figure}[t!] 
    \centering{\includegraphics[width=8.6cm]{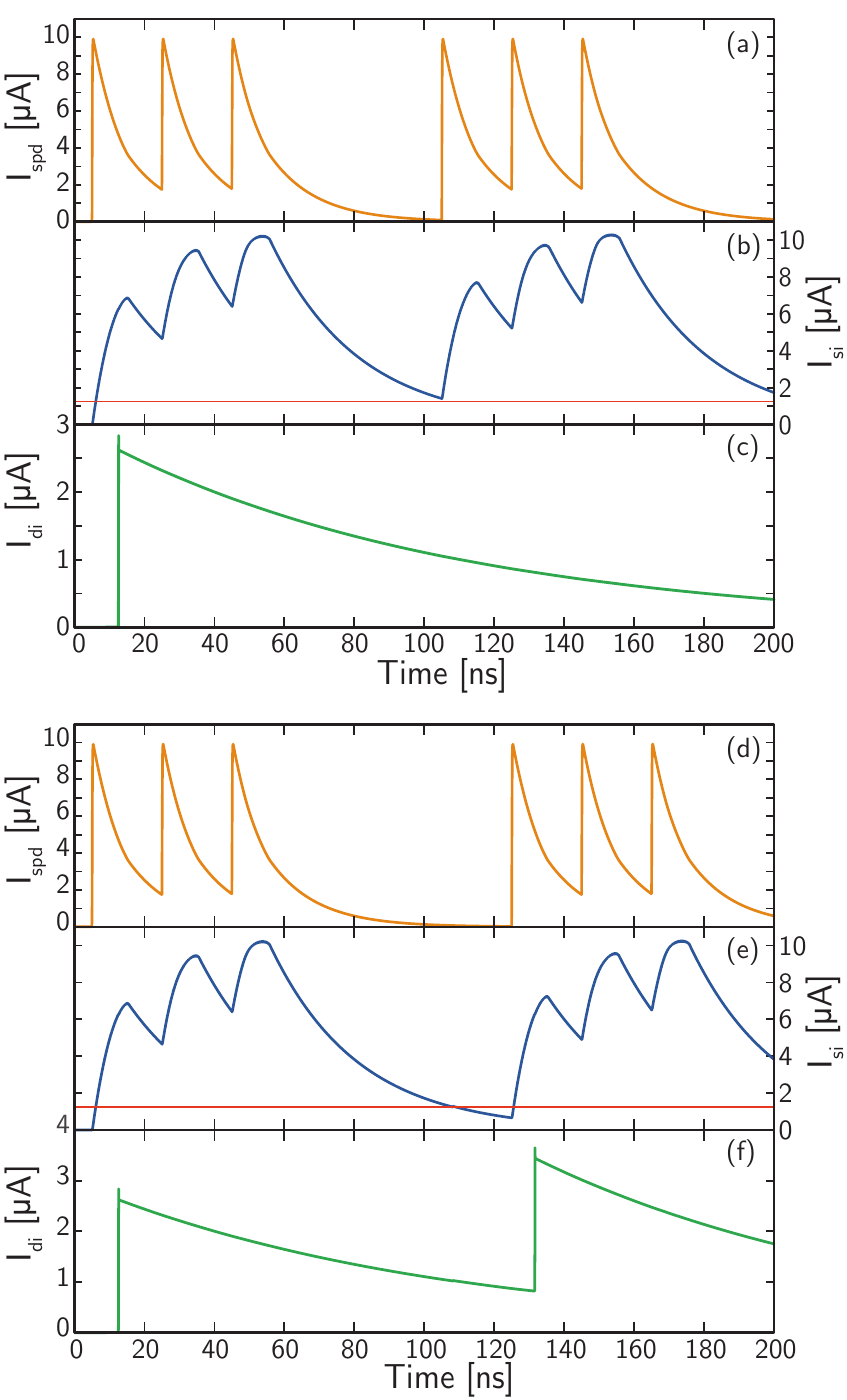}}
	\captionof{figure}{\label{fig:si_dcsfq_depressing}Short-term-depressing plasticity. With a single synapse coupled to a dendrite, the first pulse of a train can generate signal in the DI loop, and the response of the synapse is depressed for subsequent pulses until the signal in the SI loop decays below a certain level, resulting in reset. (a) Afferent activity. (b) The resulting signal in the SI loop. The red line shows the reset level, which is not quite reached before the second series of three pulses occurs. (c) The signal in the DI loop, $I_{\mathrm{di}}$. Only a single pulse enters the DI loop because the break between the pulse trains was not long enough to achieve reset. (d) Afferent activity with a slightly longer delay between the two pulse trains. (e) The signal in the SI loop, dropping briefly below the reset threshold. (f) The resulting current, $I_{\mathrm{di}}$, showing two pulses generated as the dendrite recognized these as two separate pulse trains.}
\end{figure}
While facilitating behavior effectively strengthens a synapse as a pulse train proceeds, short-term-depressing plasticity gives the opposite behavior. In an extreme form, this mechanism can be used to convey only the onset of a pulse train, while blocking subsequent spikes. To demonstrate this behavior, we consider the dendritic processing circuit in binary mode. Circuit simulations are shown in Fig.\,\ref{fig:si_dcsfq_depressing}. Consider first the upper panel, Fig.\,\ref{fig:si_dcsfq_depressing}(a-c). The current pulses from the SPD due to the afferent spike train are shown in Fig.\,\ref{fig:si_dcsfq_depressing}(a), and the resulting current in the SI loop is shown in Fig.\,\ref{fig:si_dcsfq_depressing}(b). The activity consists of two groups of three spikes. The current in the DI loop is shown in Fig.\,\ref{fig:si_dcsfq_depressing}(c). A single pulse enters the DI loop at the onset of the first spike in the train. In Fig.\,\ref{fig:si_dcsfq_depressing}(b), we have marked with a red line the value of $I_{\mathrm{si}}$ below which reset occurs in the DR loop. We see that the first spike of the second group of three occurs just before $I_{\mathrm{si}}$ drops below the reset value. The second group of pulses is not identified as a new spike train, so no additional signal is added, and $I_{\mathrm{di}}$ continues decaying with $\tau_{\mathrm{di}}$. By contrast, in the lower panel (Fig.\,\ref{fig:si_dcsfq_depressing}(d-f)), the onset of the second group of pulses occurs 20\,ns later that in the upper panel, giving the current in the SI loop (and therefore the DR loop) time to decay below the reset value. In this case, when the second group of pulses begins, it is identified as a new train, and additional signal is added to the DI loop, again in the form of a single fluxon. The reset delay can be established in hardware across a broad range of values through $\tau_{\mathrm{si}}$ and can be adjusted over a smaller range dynamically through $I_{\mathrm{de}}$. The dendritic receiving loop does not have any resistance of its own, so the current decay time constants in that loop are entirely determined by the SI loops. 

While we refer to this operation of the dendritic processing circuit as binary, the DI loop may be independently configured to store anywhere from one to many fluxons, providing the circuit as a whole with an analog representation of the number of afferent pulse trains occurring within a time period set by $\tau_{\mathrm{di}}$. If the DI loop is configured with large $\beta_{\mathrm{L}}$ and $\tau_{\mathrm{di}}$ on the order of theta time scales, the dendrite will keep track of how many gamma-frequency pulse trains have occurred, thereby keeping track of oscillations on theta time scales. Because the maximum signal level in the DI loop can be made the same as in an SI or DI loop keeping track of gamma activity, such dendritic processing is capable of representing gamma and theta information with equal weight. Alternatively, using the same circuit configuration except employing an SI loop with a time constant close to $\tau_{\mathrm{spd}}$ will cause the DI loop to receive a single fluxon each time the synapse receives a photon. In this mode of operation, the circuit achieves single-photon-to-single-fluxon transduction, converting each photon detection event to an identical, binary signal. If synaptic weighting is not required, and dendritic weights alone can suffice, the signal from a photon-detection event can immediately be converted to a single fluxon, and energy efficiency can be gained.

To summarize the operations we have investigated so far, the synaptic firing circuit on its own can accomplish rate-to-current conversion, reporting a temporal average of recent activity. By coupling the synaptic firing circuit to a dendritic processing circuit, we can construct a dendrite that generates signal only when a pulse train persists for a certain duration. We can use the same circuits with slightly different biasing configuration to construct a dendrite that generates signal only when a pulse train begins after a certain period of rest. All of these operations correspond to temporal filters performed on spike trains occurring at a single synapse. Yet an important function of dendritic processing is to identify coincidences and sequences between the activities of multiple neurons. We now consider this task.

\section{\label{sec:correlations}Detecting coincidences between neurons}
\begin{figure}[t!] 
    \centering{\includegraphics[width=8.6cm]{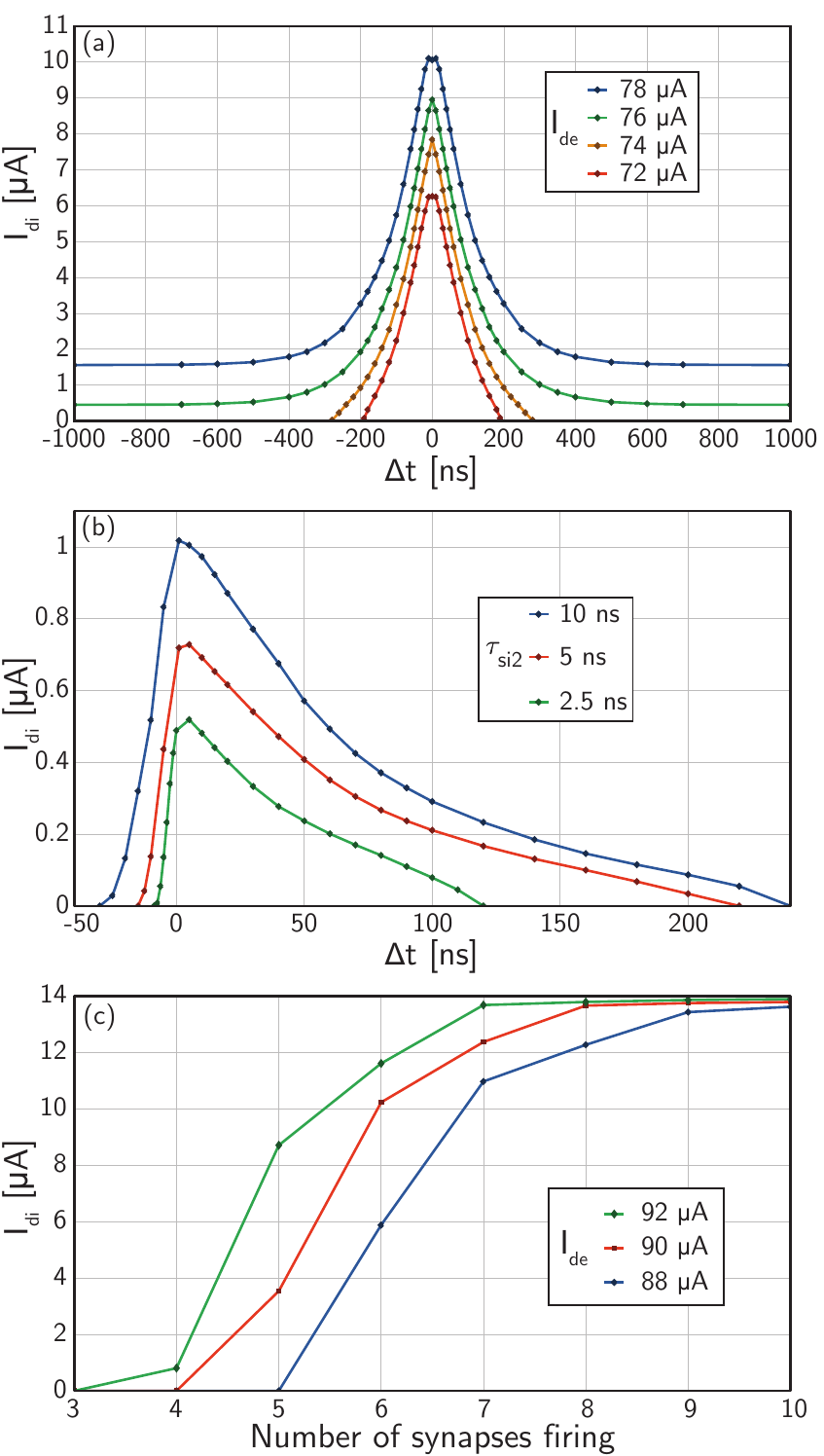}}
	\captionof{figure}{\label{fig:poly_si}Multiple synapses from different neurons coupled to a single DI loop. (a) Two synapses with the same time constant. The current induced in the DI loop ($I_{\mathrm{di}}$) is plotted as a function of the time between the two synapse events ($\Delta t$) for three values of the dendritic bias current ($I_{\mathrm{de}}$). (b) Two synapses with significantly different time constants. $I_{\mathrm{di}}$ is plotted versus $\Delta t$ for three values of the fast synaptic time constant ($\tau_{\mathrm{si2}}$). (c) Ten synapses from different neurons coupled to a single DI loop. The current generated in the DI loop is plotted as a function of the number of synapses firing simultaneously for three values of the dendritic bias current, $I_{\mathrm{de}}$.}
\end{figure}
%For the two lowest values of $I_{\mathrm{de}}$, the value added to $I_{\mathrm{di}}$ drops to zero for sufficient delay. 
The second term in a Volterra expansion of the activities of two neurons corresponds to coincidences between the two neurons \cite{geki2002}. We can use the same dendritic processing circuit of Fig.\,\ref{fig:circuits}(b) to detect coincidences, provided two SI loops are coupled to the DR loop through mutual inductors. In the simplest case, we wish to know whether two synapses have fired within a certain time period of each other. This can be achieved by giving both SI loops the same value of $\tau_{\mathrm{si}}$. The response of such a circuit is shown in Fig.\,\ref{fig:poly_si}(a), where the current induced in the DI loop is shown as a function of the time delay between the two synaptic firing events for several values of $I_{\mathrm{de}}$ with $\tau_{\mathrm{si}} =$100\,ns. For the two lower values of $I_{\mathrm{de}}$, the circuit can be thought of as an AND gate with an analog extension to the time domain: if synapse $i$ AND synapse $j$ fire within a time period set by $\tau_{\mathrm{si}}$, a signal dependent on the time difference is added to the DI loop. For larger values of $I_{\mathrm{de}}$, the circuit performs an OR operation, because for arbitrarily large $\Delta t$, the current in one SI loop alone is sufficient to switch $J_{\mathrm{df}}$ and generate some signal in the DI loop. A similar coincidence detection circuit was proposed in Ref.\,\onlinecite{sh2018_full} based on two SPDs. The advantage of the circuit presented here is that the computation occurs in the electronic domain, bringing the advantage of energy efficiency as well as the ability to perform multiple dendritic operations simultaneously through the use of fluxonic pulse splitters (Sec.\,\ref{sec:fluxonic_fanout}).

The dendritic tree may benefit from the ability to detect not just coincidences, but also the specific sequence in which synapse events occurred \cite{haah2015}. This can be achieved by breaking the symmetry between the two synapses with $\tau_{\mathrm{si1}} \gg \tau_{\mathrm{si2}}$. We consider this scenario in Fig.\,\ref{fig:poly_si}(b). Here, $\tau_{\mathrm{si1}}$ is still 100\,ns, but $\tau_{\mathrm{si2}}$ is much shorter, and we again plot the current added to the DI loop as a function of $\Delta t = t_2-t_1$, where $t_i$ is the time of a synapse event on synapse $i$. In this case, the response function is highly skewed toward $\Delta t > 0$. It is highly probable that any current induced in the DI loop is due to an event on synapse one followed by an event on synapse two. Yet with this simple design, the contribution from $\Delta t < 0$ does not vanish completely. We have plotted the response for three values of $\tau_{\mathrm{si2}}$. We see that as we decrease $\tau_{\mathrm{si2}}$, the error due to current added when $\Delta t < 0$ decreases as $\tau_{\mathrm{si2}}$ decreases. Thus, we can tighten the timing tolerance by decreasing $\tau_{\mathrm{si2}}$. With $\tau_{\mathrm{si2}} = $2.5\,ns, errors do not occur if $t_2$ is prior to $t_1$ by 8\,ns, less than the interspike interval of a gamma sequence, rendering this circuit capable of providing reliable information regarding the temporal order of activity between two synapses.

The coincidence and sequence operations of the dendritic processing circuit provide information regarding activity at two synapses. We would like to extend this to perform nonlinear operations on groups of multiple synapses. This can be straightforwardly achieved by coupling multiple synapses to a single dendrite, using the same circuits we have been discussing so far. In Fig.\,\ref{fig:poly_si}(c) we show the value of $I_{\mathrm{de}}$ resulting from a variable number of synapses firing simultaneously, with 10 total synapses coupled to a DR loop. We have chosen the circuit parameters so the bias added to $J_{\mathrm{df}}$ by a single synapse event is insufficient to exceed $I_{\mathrm{c}}$. The transfer function of the circuit is highly nonlinear, approximating a sigmoidal activation function. Thus, the current generated in the DI loop is not the sum of independent SI currents (see Ref.\,\onlinecite{geki2002}, pg. 101). The threshold number of active synapses can be set in design across a broad range, and as the three traces reveal, this number can be dynamically adjusted with $I_{\mathrm{de}}$. In both Fig.\,\ref{fig:poly_si}(a) and Fig.\ref{fig:poly_si}(c) we see that $I_{\mathrm{de}}$ can be used in a manner analogous to the synaptic bias current, pointing to the potential for reconfigurable efficacy and learning. While Fig.\,\ref{fig:poly_si}(c) only considers simultaneous synaptic activity, the true response of the dendrite would convolve the temporal responses of the constituent synapses. Similar principles to those demonstrated in Figs.\,\ref{fig:poly_si}(a) and (b) shape the net dendritic contribution.

All operations discussed thus far are excitatory. We now turn our attention to inhibition of the dendritic response.

\section{\label{sec:inhibition_and_rapid_query}Inhibition and rapid query}
The dendritic tree offers the most information to the neuron when it can be dynamically adapted into diverse functional networks. Inhibition can enable such adaptation (as well as many additional functions \cite{robu2015}) by temporarily silencing specific dendrites or entire branches of the dendritic tree. To accomplish this with the dendritic processing circuit under consideration, we couple an additional loop to the DR, except with mutual inductor of reverse coupling to oppose the bias to $J_{\mathrm{df}}$. We refer to this as an inhibitory (IH) loop, as shown in Fig.\,\ref{fig:circuits}(b). The circuit parameters can be chosen so that following a synaptic event on the inhibitory synapse no amount of activity on the excitatory synapses can drive $J_{\mathrm{df}}$ above $I_{\mathrm{c}}$. As discussed above, the AND/OR logical operations become coincidence detections when extended to the time domain \cite{stsp2015}, and when the previously considered AND circuit is augmented with an inhibitory input, the logical operation becomes AND-NOT \cite{stsp2015}.  

\begin{figure}[t!] 
    \centering{\includegraphics[width=8.6cm]{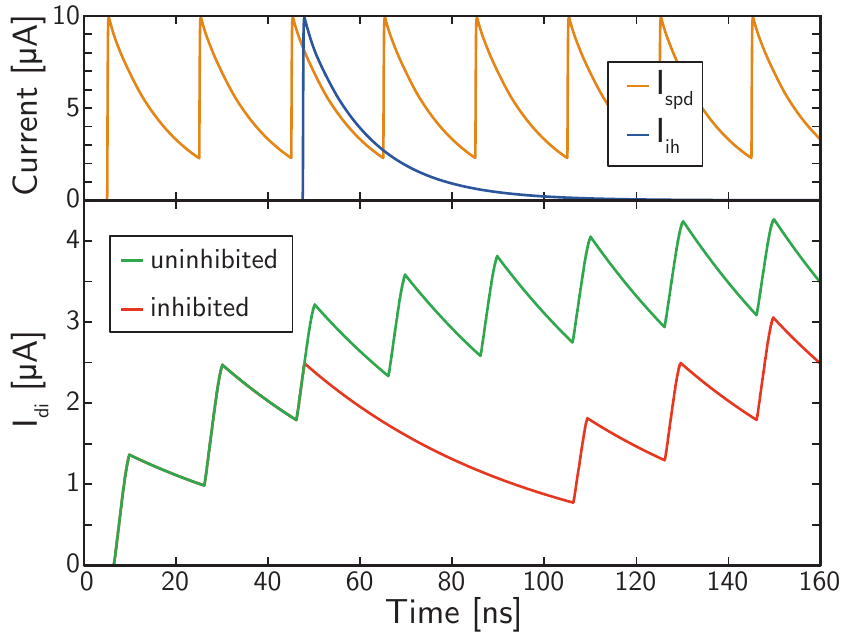}}
	\captionof{figure}{\label{fig:si_di_inhibition}The effect of inhibition. A single excitatory synapse and a single inhibitory synapse are coupled to a DR loop. The upper panel shows the signal from an afferent pulse train as well as the single inhibitory pulse. The lower panel shows the current generated in the DI loop ($I_{\mathrm{di}}$) with and without the inhibitory pulse. In this example, the circuit has been configured so that following the inhibitory pulse, no amount of activity on the excitatory synapse can drive $J_{\mathrm{df}}$ above $I_{\mathrm{c}}$, and the dendrite is completely suppressed until the signal in the IH loop has decayed.}
\end{figure}
Simulated operation of a dendrite with a single excitatory and single inhibitory synapse is shown in Fig.\,\ref{fig:si_di_inhibition}. The upper panel shows a temporal trace of excitatory activity, which consists of a pulse train at 50\,MHz. A single inhibitory synapse event occurs shortly after the third pulse of the excitatory train. The lower panel shows the current circulating in the DI loop as a function of time for cases with and without the inhibitory synapse event. Without inhibition, current is added to and decays from the DI loop, as expected. When inhibition occurs, the effect of excitation is immediately quenched. Following the inhibitory synapse event, $I_{\mathrm{di}}$ begins decaying with time constant $\tau_{\mathrm{di}}$. Inhibition decays with a completely independent time constant, $\tau_{\mathrm{ih}} = L_{\mathrm{ih}}/r_{\mathrm{ih}}$, just as all other loops discussed thus far. When the inhibitory current has decayed sufficiently, the effect of the excitatory pulse train resumes. 

The duration over which the dendrite is inhibited is controlled by $\tau_{\mathrm{ih}}$, and for the network to be rapidly adaptable under the influence of inhibition, this time constant will be as short as a gamma-range interspike interval. If inhibition is required over theta time scales, repeated activity on the inhibitory neuron can keep the dendrite suppressed. However, this may not be the most energy-efficient mode of operation. Given the circuits under consideration, we can utilize a mode of operation complimentary to inhibition. In this configuration, the mutual inductors and bias to the DR loop are chosen so that even with all afferent SI loops saturated, the current across $J_{\mathrm{df}}$ cannot exceed $I_{\mathrm{c}}$. Only when an additional, unique synapse fires does the current exceed $I_{\mathrm{c}}$. The additional synapse is designed to saturate with each synapse event and to decay rapidly with identical response to each synapse event and no synaptic weight variation. The action of this synapse is to allow $J_{\mathrm{df}}$ to sample $I_{\mathrm{dr}}$. When this synapse fires, the current generated in the DI loop provides an answer to the question, ``How much current is in the DR loop?'' We refer to neurons making synaptic connections of this type as rapid query neurons. 

\begin{figure}[t!] 
    \centering{\includegraphics[width=8.6cm]{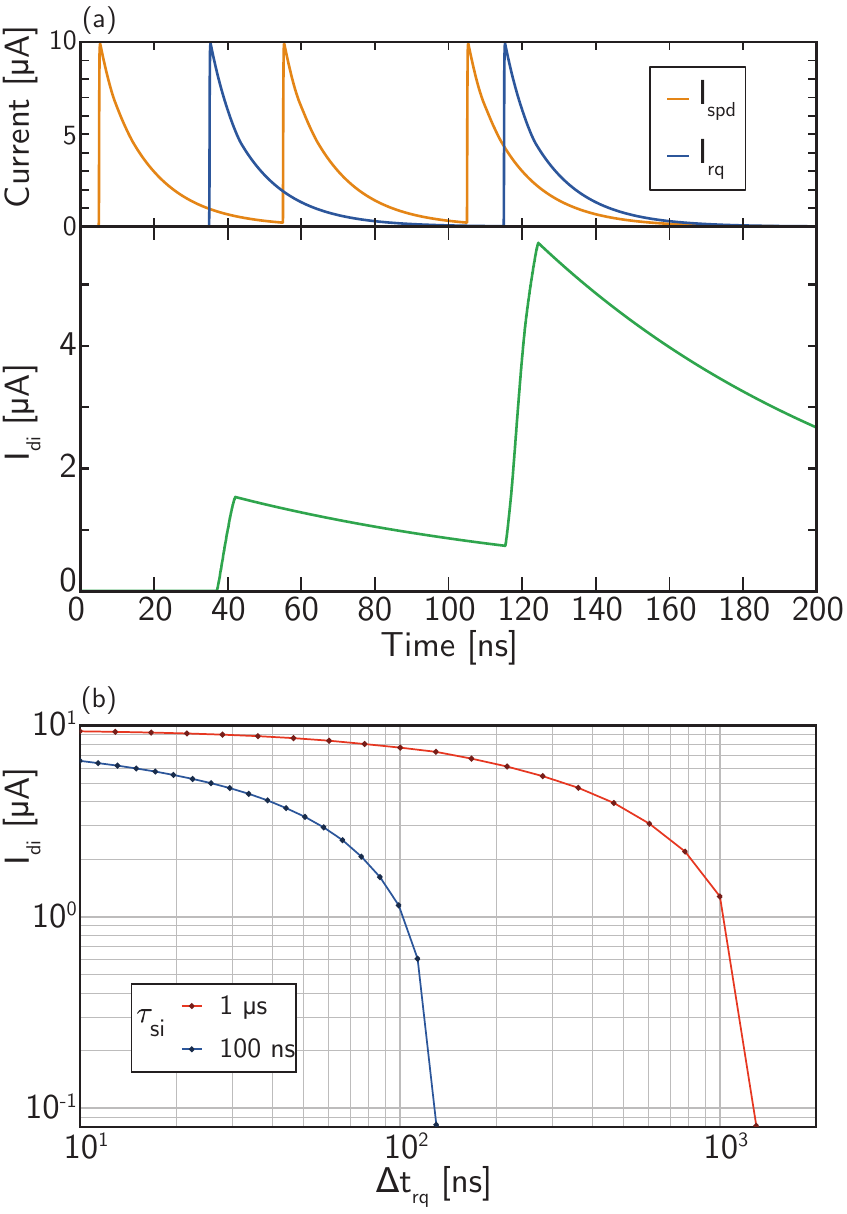}}
	\captionof{figure}{\label{fig:si_di_rapid_query}Rapid query. (a) Temporal response of a single excitatory synapse and a single rapid query synapse coupled to a DR loop. The upper panel shows the pulses resulting from photon detection events on the excitatory and rapid query synapses, while the lower panel shows the the current generated in the DI loop by the two rapid query events. (b) Systematic quantification of the result of rapid query activity following a single excitatory pulse. The current generated in the DI loop, $I_{\mathrm{di}}$, is plotted as a function of the time delay between the excitatory synapse event and the rapid query synapse event. All simulations in this figure were conducted with $\tau_{\mathrm{rq}} = 10$\,ns.}
\end{figure}
Figure \ref{fig:si_di_rapid_query} considers rapid query operation. The circuit under consideration comprises a single excitatory synapse and a single rapid query synapse coupled to a DR loop in the configuration of Fig.\,\ref{fig:circuits}(b). In the present example, three excitatory synapse events occur, as seen in the upper panel of Fig.\,\ref{fig:si_di_rapid_query}(a). Two rapid query synapse events are also shown in that panel. The first rapid query event follows the first excitatory pulse by 30\,ns, and with $\tau_{\mathrm{si}} = 20$\,ns, only a small amount of current is added to the DI loop. The second excitatory event is not followed by a rapid query event, and no current is added to DI. The third excitatory event is followed by a rapid query event with 10\,ns delay, and significantly more current is induced in the DI loop. 

The behavior of this circuit is summarized more systematically in Fig.\,\ref{fig:si_di_rapid_query}(b). Here we plot the current induced in the DI loop as a function of the time delay between the rapid query and excitatory events for two values of $\tau_{\mathrm{si}}$. We see that the signal generated by rapid query follows the exponential decay of the SI loop, thus providing an accurate mapping of $I_{\mathrm{si}}$ to $I_{\mathrm{di}}$ at the time rapid query was performed. 

We plot the exponential functions of Fig.\,\ref{fig:si_di_rapid_query}(b) on a log-log graph to emphasize that each SI loop provides information over a single time scale determined by $\tau_{\mathrm{si}}$. It would be desirable to find a means by which a memory trace may be extended across multiple time scales from a single photonic synapse event. This increased temporal dynamic range is one example of what can be achieved if electronic copies of photonic synapse events are produced. This fluxonic fan-out is the subject of the next section.

\section{\label{sec:fluxonic_fanout}Fluxonic fan-out from photonic synapses}
In neural systems using light for communication, generation and detection of photons are likely to consume the most energy. We have described several example operations that can be performed to extract information from photonic synapse events and pulse trains, and we would like to perform them all simultaneously without requiring an additional photonic synapse for each. We can straightforwardly copy fluxons with a pulse splitter, a common means of achieving fan-out of flux-quantum signals \cite{lise1991}. We can therefore simply copy the output signals from a single photonic synapse to multiple independent SI loops that can each perform different temporal filters and feed into different dendrites. We refer to these as electronic synapses, and we anticipate that each photonic synapse will feed multiple electronic synapses.

The circuit for splitting pulses is shown in Fig.\,\ref{fig:circuits}(c). A fluxon enters from the left, and when it switches the initial junction, the current of the resulting fluxon is split to two subsequent junctions. These junctions are biased such that the amount of current is sufficient to exceed $I_{\mathrm{c}}$, thus producing fluxons at both junctions with restored signal level. For the application at hand, the splitter of Fig.\,\ref{fig:circuits}(c) can be placed following $J_{\mathrm{jtl}}$ in Fig.\,\ref{fig:circuits}(a) or Fig.\,\ref{fig:circuits}(b). Thus, signals produced by synapses or dendrites can be copied and processed independently to extract distinct information through multiple temporal filters and logical operations. The circuit of Fig.\,\ref{fig:circuits}(c) achieves direct one-to-two fan-out. If a greater number of copies is desired, the same circuit can be repeated in a tree. The limits of this fan out will depend on one's tolerance for circuit complexity. We speculate that in mature systems, a given photonic synapse may split to as many as 10 electronic synapses. 

\begin{figure}[t!] 
    \centering{\includegraphics[width=8.6cm]{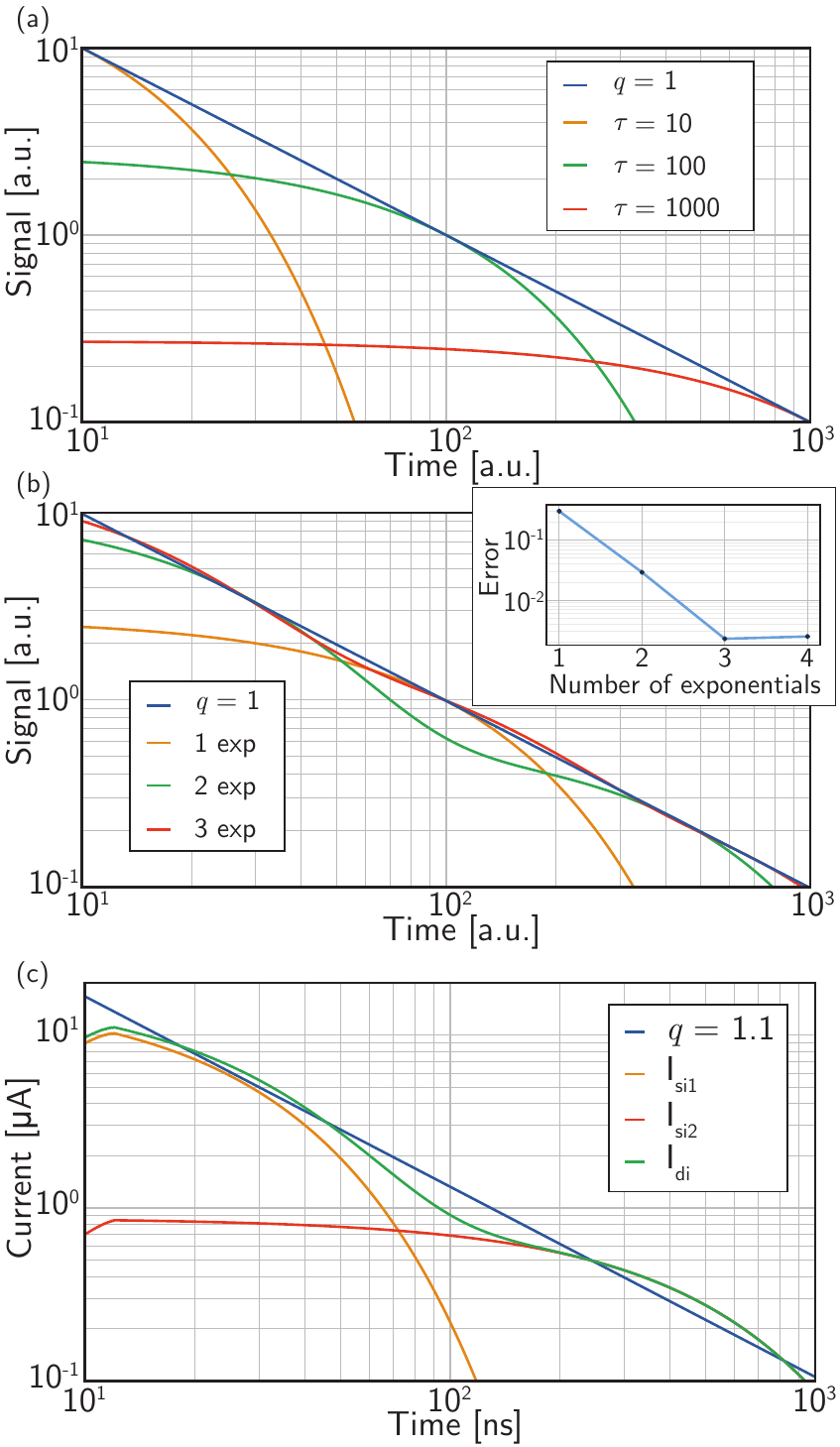}}
	\captionof{figure}{\label{fig:power_law_time_decay}Approximation of power-law temporal decay. (a) Illustration of the differences between power-law and exponential decay on a log-log plot. The functions plotted are $f(t) \propto t^{-q}$ and $g(t) \propto e^{-t/\tau}$, referred to generally as ``Signal''. (b) Approximating power-law decay through the superposition of multiple exponentials. A power law function with unity exponent is shown, as are approximations composed of one to three exponentials. Amplitudes and time constants were adjusted for best fit, and convergence is shown in the inset. (c) Approximating power-law decay with two SI loops coupled to a common DR loop. The time constants and mutual inductances have been chosen to approximate a power law using the same fitting algorithm that generated (b).}
\end{figure}
As a simple example of the utility of pulse splitting, we consider one photonic synapse feeding into two electronic synapses with different time constants. Figure \ref{fig:power_law_time_decay}(a) summarizes the motivation. Instead of retaining a memory trace of a synapse event over only a single temporal scale, as occurs in a single SI loop with exponential decay, we would prefer a signal with a power-law decay, so that information across temporal scales can be accessed. In Fig.\,\ref{fig:power_law_time_decay}(a) we compare $f(t) \propto t^{-q}$ to $g(t) \propto e^{-t/\tau}$ for three values of $\tau$. The smallest value of $\tau$ provides no information past its cutoff, and the signal from the largest value of $\tau$ is nearly constant initially across more than an order of magnitude. The middle value gives a poor representation at the start and the end. Figure \ref{fig:power_law_time_decay}(b) shows that we can obtain a suitable approximation to the power law function by superposing a small number of exponentials \cite{be2007}. Here we represent a power law with unity exponent, mapping two orders of magnitude in time to two order of magnitude in signal. Convergence is shown in the inset. The error is improved by an order of magnitude when using two exponentials instead of one, and there is little advantage to using more than three for this task. 

We implement this principle with the circuits under consideration by copying the signal from a photonic synapse to two electronic synapses coupled to a common passive superconducting loop via mutual inductors. We choose the time constants and couplings of the two SI loops to approximate the fitting technique employed to produce Fig.\,\ref{fig:power_law_time_decay}(b). Figure \ref{fig:power_law_time_decay}(c) shows the current in each of the SI loops as well as the common output loop. A power law with $q = 1.1$ is shown for comparison. This power-law temporal extension can be used in conjunction with many of the other operations discussed thus far, with the objective to use cheap fluxonic operations to extend the memory trace of expensive photonic activity across extra orders of magnitude in time. Such operation performs a power-law mapping of a temporal signal to the dynamic range of the firing junction, and allows a single dendrite to retain and access information regarding both gamma and theta frequencies. 

This example of using pulse splitting to access broader time spans is a straightforward extension of the behavior of a single SI loop. Additional functionality can be envisioned by combining pulse splitting with many of the functions discussed in this paper. Most importantly, by copying the output from a photonic synapse, each of the operations discussed here can be performed concurrently. With a single photon, the dendritic tree can be provided with information regarding the synapse's average firing rate across multiple temporal scales; the time since the last synaptic firing; various quantities regarding initiation and duration of pulse trains; coincidences and sequences with synapses from multiple other neurons; and inhibition and rapid query applied independently to each of these pieces of information.

\section{\label{sec:discussion}Summary and discussion}
We have described several synaptic and dendritic operations achieved with Josephson junctions and mutual inductors. These include various logical operations, temporal filters, and nonlinear transfer functions applied to one or more synapses. The operations performed here are all accomplished with configurations of the building blocks shown in Fig.\,\ref{fig:circuits}. We envision the dendritic tree to be comprised of a complex network of synapses and dendrites performing a multitude of computations on signals that fan in from photonic synapses, traverse the dendritic tree, and feed the neuron's final thresholding compartment, which triggers the production of light. A network will comprise many neurons, and each neuron is itself a network. We have described the dynamic functional adaptation of the dendritic network through inhibition and rapid query. Inhibitory activity nullifies targeted portions of the tree, while rapid query obtains local fragments of information and passes them along the tree. We have also described how electronic copies of photonic synapse events can enable several of these operations to be performed with the information from the detection of a photon. 

This work provides additional support for the hypothesis that superconducting computation is complimentary to photonic communication for achieving large-scale neural systems. While photons can achieve fan-out, they lack the required nonlinearities required for computation, especially at the low light levels required for energy efficiency. Further, photons cannot be made to sit still for memory retention. Additionally, generating photons is more expensive than generating fluxons, and therefore only the minimum number of photons required for communication should be generated. Superconducting circuits are complimentary to photonic circuits in these regards. The proposed hardware aspires to achieve greater than one-to-one-thousand fan-out in the photonic domain from each neuron to its thousands of connections \cite{sh2018_ICRC}, and subsequently, at each neuronal terminal, the hardware aspires to achieve an additional factor of roughly one-to-ten fan-out in the electronic domain, providing each receiving neuron with the capability of analyzing much more information about synaptic activity. Fan-in is envisioned to occur in the electronic domain as the dendritic tree computes and feeds its signals into the neuron cell body, ultimately resulting in a binary decision of whether or not to fire. Superconducting-nanowire single-photon detectors enable binary communication in that the response is nearly identical whether one or more photons are detected, and all computations\textemdash including synaptic weighting, nonlinear processing, and temporal integration\textemdash occur in superconducting electronic circuits with sub-nanosecond response times, native nonlinearities, and the potential for signal retention with no dissipation.
 
In mature superconducting optoelectronic circuits, we would like the energy expended on light production, photon detection, and fluxonic processing to be roughly equal. Production of a fluxon requires $E_{\mathrm{j}} = I_{\mathrm{c}}\Phi_0/2\pi = 1.3\times 10^{-20}$\,J for the junctions considered here, while production of a photon requires $E_{\mathrm{p}} = h\nu/\eta = \frac{1.6\times 10^{-19}}{\eta}$\,J, where $\eta$ is the photon production efficiency, and we consider operation at $\lambda = 1.22$\,\textmu m \cite{buch2017}. Light generation is expensive because $\eta$ is unlikely to ever exceed 0.1 and may be limited to 0.01 or worse. Likewise, photon detection requires $E_{\mathrm{d}} = L_{\mathrm{spd}}I_{\mathrm{spd}}^2/2 = 1.3\times 10^{-17}$\,J for the superconducting-nanowire single-photon detector designs presented here. Due to the requirement of engineering reset dynamics in the detector, $L_{\mathrm{spd}}$ cannot be reduced below a certain value without decreasing the normal-state resistance of $J_{\mathrm{sf}}$, which requires increasing $I_{\mathrm{c}}$, which increases $E_{\mathrm{j}}$. Similarly, $I_{\mathrm{spd}}$ cannot decrease without using either junctions with smaller $I_{\mathrm{c}}$ or operating them in a noisy regime with bias close to $I_{\mathrm{c}}$, and would result in reduction of the dynamic range of the synaptic weight. This space of trade-offs is complex, and we make no attempt to identify the optimum in this work. We simply note that if $\eta = 0.01$, $I_{\mathrm{spd}} = 10$\,\textmu A, $L_{\mathrm{spd}} = 250$\,nH, $I_{\mathrm{c}} = 40$\,\textmu A, and full analog processing of each synapse event generates $10^3$ fluxons on average, then light generation, detection, and fluxonic processing each contribute roughly equally to energy consumption. Full optoelectronic integration with few-photon binary communication and superconducting electronic analog computation offers a route to balance the energy budget while enabling the requisite communication and repertoire of computational functions for large-scale artificial cognitive systems.

One emphasis in this work has been on the interaction of inhibitory and rapid query neurons with dendrites to enable diverse functional networks. Inhibition is central to neural computation \cite{robu2015}, with a key role being the formation and synchronization of adaptive neuronal modules that operate as task-specific processors \cite{vala2001,sase2001}. With inhibition, branches of the dendritic tree are functionally responsive by default and are selectively silenced by inhibitory synapse events. Inhibition can lead to synchronization by opening brief temporal windows when groups of neurons can fire \cite{bu2006}. With rapid query, branches of the dendritic tree are silent by default and are only functionally connected if rapid query synapse events occur. If the information in a given dendrite need not be accessed regularly, rapid query will be more energy efficient than continually performing inhibition. Like inhibition, rapid query may be useful for inducing synchronization. We do not propose rapid query instead of inhibition, but rather in addition. Both inhibition and rapid query may be leveraged to enable sub-threshold oscillations to be sampled only when required by the network, as occurs in biological neural systems to direct attention and amplify relevant information \cite{enfr2001}. We posit the utility of a dedicated class of rapid query neurons in superconducting optoelectronic networks even though, to our knowledge, there is no such class of neurons in the biological domain. This may be due to a computational inadequacy of rapid query that we have overlooked, or it may be that the circuits under consideration are more amenable to such a mode of operation, which requires a degree of control over competing circuit parameters. There are dozens of different, specialized neurons in the mammalian brain, with multiple types of inhibitiory neurons playing specific roles \cite{robu2015,bu2006}. Superconducting optoelectronic networks take significant inspiration from the brain, but hardware discrepancies will inevitably lead to deviations in computation. Perhaps rapid query neurons are one such departure.

There are multiple possible extensions of the functions considered here as well as further details to be considered. XOR may be achieved with pulse splitting and lateral inhibition between dendrites. We have only considered binary inhibition, but weaker or multiple IH loops could be coupled to a DR loop to achieve partial inhibition. The neural operations considered here tend toward analog operation of the superconducting circuits, and we have presented circuits capable of representing signals with eight to 10 bits of resolution based on the $\beta_{\mathrm{L}}$ values chosen for the integration loops. However, this resolution is only available if noise is sufficiently low, so further investigation is required to determine a suitable tradeoff between loop inductance, signal resolution, and operating temperature. Future work may find different optimal values for different operations, and an improved balance between information capacity and hardware demands might be discovered. We have primarily considered signal storage loops with retention times on the order of what we suspect will be the gamma and theta frequencies of the system, but further research may find advantages of retaining fading memories for much longer than this or may reveal that even theta retention times are gratuitous. 

In several instances we have indicated that the dendritic bias current $I_{\mathrm{de}}$ can be used to adjust circuit operation, pointing to a means of achieving learning and plasticity between synapses and dendrites \cite{haah2015} or between two dendrites. This subject deserves further investigation, but at present we simply note that similar circuits used for spike-timing-dependent plasticity in Ref.\,\onlinecite{sh2018_full} can be used to implement such activity-based weight update functions. 

We have only considered the first layer of dendritic hierarchy, but the same dendritic building block of Fig.\,\ref{fig:circuits}(b) can be tiled essentially arbitrarily. The depth of this tree is enabled by the logic-level restoration occurring in the basic circuit. Design of the DI loop is independent of the DR loop, and regardless of the configuration of the inputs to the DR loop, as long as threshold can be reached, flux can be added to the DR loop, and a restored current level can be attained with as few as one fluxon. In this work, that current level is around 10\,\textmu A, but it could be designed to be higher or lower as needed. This logic-level restoration enables a many-compartment dendritic tree to be as deep as needed for the desired information processing, pointing to numerous theoretical questions. At the base of the tree is the soma, or cell body. The soma receives signals just as any of the other dendrites, but its output feeds into an amplifier chain that leads to the production of light \cite{sh2018_full}. Because nanowire single-photon detectors have a binary response, each neuron-to-synapse communication event also results in logic-level restoration, but between neurons and synapses rather than dendrites.

Beyond specifics related to the superconducting optoelectronic hardware implementation, this work touches on important theoretical questions regarding neural information. We have based circuit designs around the hypothesis that incorporating significant dendritic structure beyond the point-neuron model is important for neural processing. Quantification of dendritic information processing is difficult in biological experiments due to the length scales involved, the sensitivity of the neurons and dendrites under study, and the inability to design or control the circuits being investigated. The circuits presented here can be precisely designed, fabricated, manipulated, and measured, potentially leading to traction on theoretical models of dendritic processing. The goal of the dendritic tree is to provide as much information as possible about the temporal activity on a neuron's afferent synapses. Proper design will maximize knowledge in the dendritic tree and the arbor's ability to communicate that information to the cell body. Versatile hardware implementations of neurons with various dendritic processing capabilities may serve to elucidate the important functions of dendrites in biological and artificial neural systems.

\vspace{2em}
This is a contribution of NIST, an agency of the US government, not subject to copyright.

\bibliographystyle{unsrt}
\bibliography{fluxonic_processing_of_photonic_synapse_events}

\end{document}